OPEN

# Data Descriptor: Computed tomography data collection of the complete human mandible and valid clinical ground truth models

Jürgen Wallner[1,2], Irene Mischak[3] & Jan Egger[1,2,4]

Image-based algorithmic software segmentation is an increasingly important topic in many medical fields. Algorithmic segmentation is used for medical three-dimensional visualization, diagnosis or treatment support, especially in complex medical cases. However, accessible medical databases are limited, and valid medical ground truth databases for the evaluation of algorithms are rare and usually comprise only a few images. Inaccuracy or invalidity of medical ground truth data and image-based artefacts also limit the creation of such databases, which is especially relevant for CT data sets of the maxillomandibular complex. This contribution provides a unique and accessible data set of the complete mandible, including 20 valid ground truth segmentation models originating from 10 CT scans from clinical practice without artefacts or faulty slices. From each CT scan, two 3D ground truth models were created by clinical experts through independent manual slice-by-slice segmentation, and the models were statistically compared to prove their validity. These data could be used to conduct serial image studies of the human mandible, evaluating segmentation algorithms and developing adequate image tools.

| **Design Type(s)** | modeling and simulation objective • image processing objective |
|---|---|
| **Measurement Type(s)** | 3D structure determination assay |
| **Technology Type(s)** | computed tomography |
| **Factor Type(s)** | biological sex • age |
| **Sample Characteristic(s)** | Homo sapiens • mandible |

[1]Department of Oral and Maxillofacial Surgery, Medical University of Graz, Auenbruggerplatz 5/1, 8036 Graz, Austria. [2]Computer Algorithms for Medicine Laboratory, Graz, Austria. [3]University Clinic of Dental Medicine and Oral Health, Medical University of Graz, Billrothgasse 4, 8010 Graz, Austria. [4]Institute for Computer Graphics and Vision, Graz University of Technology, Inffeldgasse 16c/II, 8010 Graz, Austria. Correspondence and requests for materials should be addressed to J.W. (email: j.wallner@medunigraz.at)

## Background & summary

Software programs for processing medical image data have become a central tool in many medical disciplines[1] and are subject to continuous ongoing technological developments. This is especially true because the total number of data sets has increased with the wider application of image-processing techniques in most clinical centres, the improving resolution of new image scanner generations has allowed medical image data sets to more accurately record anatomy, and the time required to computationally reconstruct image data has been reduced[2]. Clinically, the produced data can be used for image-based 3D visualizations of complex anatomical structures to support medical diagnosis and/or therapeutic planning procedures[3,4]. In the field of craniomaxillofacial surgery, this approach is used for complex facial bone trauma, trauma-related or congenital craniomaxillofacial deformities and orthognathic surgery[5,6].

Therefore, improvements in the software used for 3D reconstruction and volume rendering generated by image-based segmentation algorithms are needed[7,8]. Image-based segmentation methods are algorithmic software tools used for radiological image data. These image data mostly originate from computed tomography (CT) data sets[9–12].

In the craniomaxillofacial (or head and neck) complex, automated and/or semi-automated segmentation techniques have previously been used for the lower jaw to investigate their use in clinical diagnosis and treatment support[13–18].

For example, previously published works in this field include a CT data collection of lower jaws from sharks[19] and a collection of CT data for the Reference Image Database to Evaluate Therapy Response (RIDER)[20]. However, accessible medical human CT libraries, especially clinically relevant ground truth databases (a gold standard database for segmentation programs/algorithms that is similar to a control group for image-based segmentation programs/algorithms to assess segmentation accuracy/quality), for the evaluation of image-based segmentation approaches or software processes are limited. In particular, fully segmented 3D ground truth models of complete human anatomical bone structures are lacking. Therefore, the authors of this work are currently investigating multiple image-based open-source software segmentation approaches for such ground truth models[21]. Such CT data set collections and ground truth segmentations of the same anatomical structure provide resources for conducting serial image studies of the human mandible, evaluating segmentation algorithms and developing adequate image tools to assess clinical or scientific use.

Valid ground truth models of complete anatomical structures are needed to evaluate the practical use of segmentation algorithms, which are used in many clinical centres for 3D visualization or in diagnosis and treatment support in complex medical cases for both clinical and research purposes. Furthermore, such ground truth models can be used in medical image segmentation challenges, such as the Multimodal Brain Tumor Segmentation Challenge (BraTS) (https://braintumorsegmentation.org/).

This contribution provides a description of valid fully segmented ground truth CT data sets of the complete anatomy of the mandibular bone without artefacts or missing slices (Data Citation 1). The ground truth data were created by medical specialists and statistically tested for validation. These virtual 3D ground truth models can be applied to compare and analyse 3D visualizations, as well as the functional stability or accuracy of image-based segmentation algorithms in medicine. These uses may include investigations and testing of commercial and conventional segmentation approaches or other current and future image-based processing techniques.

Moreover, valid 3D ground truth data of complete anatomical structures provide a visual resource for outreach, education and/or further research purposes. This possibility represents a novel application of these data. For example, these 3D models can be used for the further development of segmentation approaches, for physical reproduction using rapid prototyping or for integration into 3D PDF documents as interactive figures. Moreover, the data collection represents an objectively created, valid and homogeneous sample collective that can be used as a representative control group in comparative assessments. This point is especially relevant since the ground truth data used for this collection originate from routine clinical practice.

Some data in this collection were used in part in previously published works[17,21–23], for the testing of an open-source segmentation algorithm[17,21], for a deep learning network[22] and for the evaluation of a software tool for computer-aided planning of miniplate positioning[23]. The written description of the data, the data selection and the data assessment method in the present work therefore partly overlap with the data description, selection and assessment methods in these publications[17,21,22]. Some of the data presented herein had to be previously released and described within a previous publication for reproducibility reasons related to the testing procedure because of the journal's requirements[17].

However, the data description presented in this manuscript has additional value because it includes complete data descriptions and data parameters, detailed results and the full data sets for a human mandibular CT library. In contrast to other works in which slices with artefacts were excluded from the segmentation process[18], leading to incomplete data sets, the new value of the data description provided with this manuscript is that it includes complete segmented valid ground truth data and 3D masks of the mandible without artefacts or missing slices.

| Mandible No. | Number/Range of Slices | | In-plane | | Scanner | Dose of Scan (kV) | Slice Thickness (mm) | Slice Increment/ Space (mm) | Scan Exposure (mAs) | Age | Sex (f/m) | Body Weight (kg) | Body Height (cm) |
|---|---|---|---|---|---|---|---|---|---|---|---|---|---|
| | Overall | Lower Jawbone | Voxels | Resolution (mm²) | | | | | | | | | |
| 1 | 217 | 22–111 | 512 × 512 | 0.428 × 0.428 | Siemens/Sensation 64 | 120 | 1 | 0.6 | 320 | 55 | m | 86 | 178 |
| 2 | 154 | 9–100 | 512 × 512 | 0.383 × 0.383 | Siemens/Sensation 64 | 120 | 1 | 0.6 | 293 | 68 | m | 75 | 174 |
| 3 | 136 | 34–72 | 512 × 512 | 0.42 × 0.42 | Siemens/Sensation 64 | 120 | 2 | 0.6 | 283 | 65 | m | 96 | 190 |
| 4 | 141 | 45–83 | 512 × 512 | 0.532 × 0.532 | Siemens/Sensation 64 | 120 | 1.99 | 0.6 | 320 | 52 | f | 67 | 176 |
| 5 | 232 | 10–107 | 512 × 512 | 0.51 × 0.51 | Siemens/Sensation 64 | 120 | 1 | 0.6 | 310 | 59 | m | 91 | 184 |
| 6 | 168 | 8–73 | 512 × 512 | 0.461 × 0.461 | Siemens/Sensation 64 | 120 | 1.5 | 0.6 | 241 | 67 | f | 71 | 168 |
| 7 | 200 | 4–76 | 512 × 512 | 0.465 × 0.465 | Siemens/Sensation 64 | 120 | 1 | 0.6 | 380 | 48 | m | 87 | 180 |
| 8 | 110 | 0–48 | 512 × 512 | 0.451 × 0.451 | Siemens/Sensation 64 | 120 | 2 | 0.6 | 205 | 71 | f | 76 | 181 |
| 9 | 154 | 9–143 | 512 × 512 | 0.383 × 0.383 | Siemens/Sensation 64 | 120 | 1 | 0.6 | 271 | 75 | f | 64 | 172 |
| 10 | 298 | 5–143 | 512 × 512 | 0.471 × 0.471 | Siemens/Sensation 64 | 120 | 0.75 | 0.6 | 232 | 74 | f | 77 | 169 |
| Mean | 181 | X | 512 × 512 | X | X | 120 | 1.3 | 0.6 | 285.5 | 63.4 | X | 79 | 177.2 |
| SD | 56 | X | 0 | X | X | 0 | 0.4 | 0 | 51.1 | 9.4 | X | 10.6 | 6.9 |
| Min | 110 | X | X | X | X | X | 1 | X | 205 | 52 | X | 64 | 169 |
| Max | 298 | X | X | X | X | X | 1.9 | X | 380 | 75 | X | 96 | 190 |

Table 1. **Descriptive overview of the included cases and the detailed scan parameters for the whole data set.** mm: Millimetres, kV: Kilovolts, mAs: Milliampere-seconds, f: Female, m: Male, kg: Kilograms, cm: Centimetres.

## Methods

This study was approved by the ethics committee of the Medical University of Graz, Austria (EK-29-143 ex 16/17, Medical University of Graz, Austria), and informed consent was obtained from all subjects.

In this investigation, the human lower jaw was selected as the anatomical structure of choice because it is the largest and strongest bone in the craniomaxillofacial complex[24]. Furthermore, the lower jaw is of high clinical relevance because it has the highest occurrence of facial fractures in the craniomaxillofacial field, representing approximately 40% of such fractures, and is therefore often involved in trauma injury[25]. Clinically, the lower jaw is frequently involved in time-consuming trauma cases in which computer-assisted image-based segmentation algorithms are used to support diagnosis and treatment procedures[13–18].

### Data selection

Forty-five original CT data sets of the craniomaxillofacial complex were provided as DICOM files and collected during routine clinical practice in the Department of Oral and Maxillofacial Surgery at the Medical University of Graz in Austria. Only high-resolution data sets (512 × 512) with slices not exceeding a spacing of 2.0 mm with a 0.5 × 0.5 mm² pixel size that provided physiological, complete mandibular bone structures without teeth were included in the selection process[17]. Furthermore, no distinction was made between atrophic and non-atrophic mandibular bones; as this condition occurs physiologically, both types of bone were included during the selection process[21].

However, incomplete data sets consisting of mandibular structures altered by iatrogenic or pathological factors or fractured mandibles, as well as data sets showing osteosynthesis materials in the lower jaw, were excluded from this trial. These data sets were excluded to avoid an inhomogeneous collection of data sets. To ensure the creation of high-quality ground truth data and a homogeneous ground truth model collection, only complete physiological data sets with clear bone contours and anatomical structures without artefacts were used, according to the mentioned inclusion criteria[17].

All data sets were acquired within a twelve-month period (between 2013 and 2017). According to the inclusion criteria, 20 CT data sets were selected; 25 data sets were excluded during the selection process in routine clinical practice for diagnosis and treatment reasons. From these 20 CT data sets, 10 data sets (n = 10; 6 male, 4 female) were further selected to form an experimental segmentation group for 3D ground truth volume creation[17,21,22]. The CT scans were performed with a Siemens Somatom Sensation 64 medical scanner using a standardized scanning protocol and a B30f reconstruction filter. Table 1 describes the details and the scanning parameters and shows the physiological information for the selected data.

### 3D ground truth segmentation

The segmentation group ultimately consisted of 10 comprehensively chosen data sets. These image series were imported into the medical platform MeVisLab (https://www.mevislab.de/). To create the ground truth volumes, a manual slice-by-slice segmentation of the randomly selected lower jaw data sets was carried out in MeVisLab by two clinical experts, specialized maxillofacial surgeons A and B, who both had the same level of experience in segmentation. Additionally, one specialized radiologist supported the ground truth segmentation process in the background. More precisely, each data set was segmented manually by clinical experts A and B through the manual outlining of bone structures in each slice to create two independent ground truth schemes (ground truth A and B) for each data set[17,21] (Fig. 1).

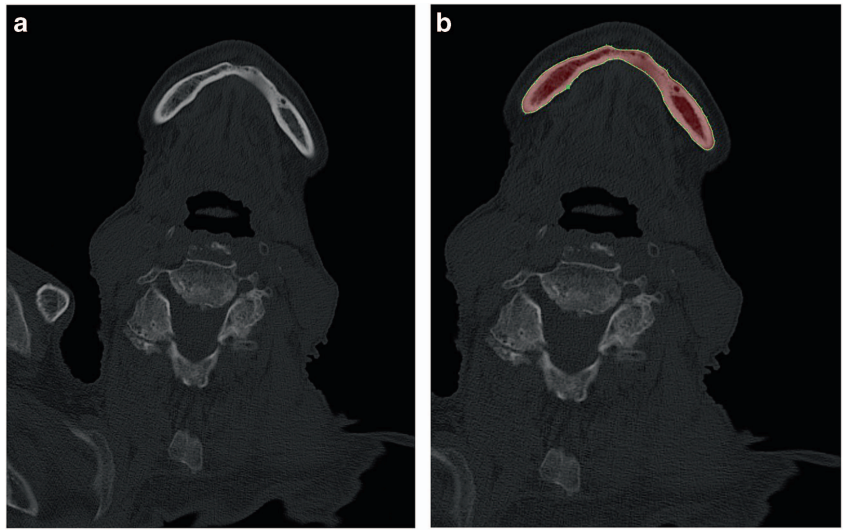

**Figure 1. Ground truth segmentation.** Each data set was segmented manually by two clinical experts, A and B, through the manual outlining of bone structures in each slice (red field). (**a**) Original CT image slice. (**b**) Manual ground truth segmentation of clinical expert A (red field).

All selected CT data sets were completely anonymized and converted from DICOM files into single Nearly raw raster data (NRRD) files. Segmentation was performed manually slice by slice for the creation of the ground truth data (MeVisLab 2.5.2., software, MeVis Medical Solutions AG, Fraunhofer Institut, Bremen, Germany)[26–29]. According to the function of the MeVisLab platform, a modular software network was integrated into the platform to outline the selected CT data sets in axial directions[17] without any algorithmic support to avoid distorting the ground truth results (Fig. 2). The selected data sets were loaded into the platforms and were successively completely segmented according to their anatomy. The segmentations were saved as contour segmentation objects (CSO) files and anonymized 3D binary masks in NRRD. After the segmentation processes were finished, 10 ground truth segmentations from clinical expert A and 10 ground truth data sets from clinical expert B were performed[17,21]. Although 10 data sets were used for the segmentation, 20 3D ground truth models are provided, due to the double segmentation (A, B) of each dataset (n = 20).

The data sets were compared by defined parameters, as follows: Ground truth A vs. Ground truth B.

### Assessment criteria

After ground truth segmentation, the created 3D models were tested for validity. The accuracy and congruence of the segmented volumes were assessed to identify any systematic inaccuracy or observer-dependent subjectivity due to the manual creation of the ground truth data. This process was performed by a comparative assessment of the two ground truth models for each data set (n = 20) using defined parameters: 1) the Dice score coefficient (DSC)[30,31], 2) the Hausdorff distance (HD)[32], 3) the segmentation volume and 4) the number of voxels (voxel units). These parameters are commonly used objective standard sizes in the evaluation of various techniques for volume and image rendering[28,29].

In more detail, the *DSC* is the agreement between two binary volumes and is calculated as follows. It measures the relative volume overlap between *R1* and *R2*, where *R1* and *R2* are the binary masks from two segmentations. *V(·)* is the volume (e.g., in mm$^3$) of voxels inside a binary mask, obtained by counting the number of voxels, then multiplying that value by the voxel size (Equation (1)).

$$DSC = \frac{2 \cdot V(R1 \cap R2)}{V(R1) + V(R2)} \quad (1)$$

The *HD* between two binary volumes is defined by the *Euclidean* distance between the boundary voxels of the masks. Given again the sets *R1* and *R2* of two segmentations that consist of the points that correspond to the centres of the segmentation mask boundary voxels in the two images, the directed *HD h(A,R)* is defined as the minimum *Euclidean* distance from any of the points in the first set to the second set, and the HD between the two sets $H(A,R)$ is the maximum of these distances[33] (Equation (2)).

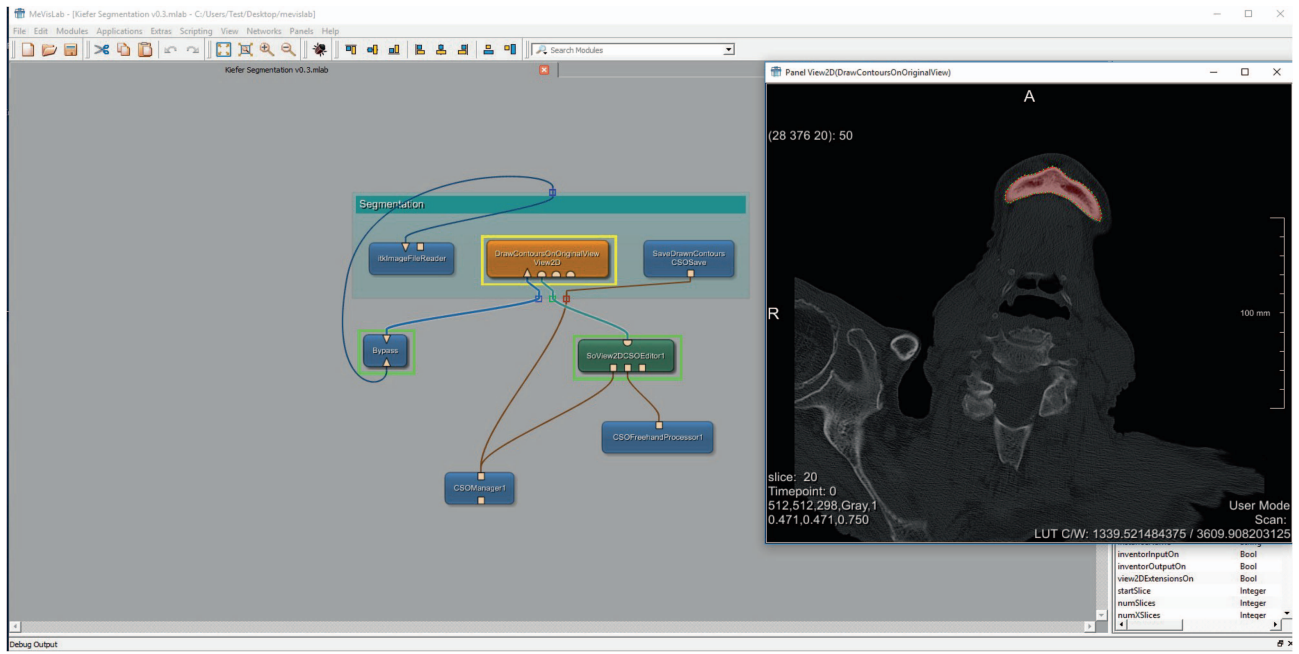

**Figure 2. Segmentation workflow.** A modular framework integrated in the software platform MeVisLab was used to outline the mandibular bone structures in the CT data sets in axial directions. The segmentation was performed without any algorithmic support to avoid distorting the ground truth results.

$$h(A, R) = \max_{a \in R}(d(a, R)), \text{ where } d(a, R) = \min_{r \in R} \|a - r\|$$
$$H(A, R) = \max(h(A, R), h(R, A)) \quad (2)$$

The measurements were then directly compared to each other (A vs. B)[17,21] to prove the validity of the models and to ensure objective ground truth models and 3D masks.

### Statistical methods
Descriptive statistical calculations were used to summarize the measurements made from the 3D models, including minimum values, maximum values, mean values and standard deviations for each of the five parameters defined above. Analytical statistical methods consisted of the calculation of paired t-tests (p) to confirm statistical validity values, the calculation of Pearson's product-moment correlation coefficient (r)[34–36] and regression analysis, including regression lines through the origin. These tests were performed to determine if there was a statistically significant difference between the measurements from each of the two sets of models (A vs B). Differences in values were calculated between the ground truth data (A, B). P-values less than 0.05 ($p < 0.05$) were considered to be significant. All statistical calculations were performed using the statistical software SPSS 20.0 (IBM SPSS Statistics; www.spss.com).

### Data records
The data for this manuscript have been stored in a figshare repository (Data Citation 1). The data sets are 10 completely anonymized CT scans of the human craniomaxillofacial complex that originally consisted of tomographic images (slices) in DICOM format from routine clinical practice. The data sets all include complete physiological mandibles, which are partly atrophic due to patient age and missing teeth but are not altered by any pathology. The corresponding mandible contours (mandibular ground truth segmentations) are provided within this data collection (Data Citation 1). To anonymize the data (visual de-identification), a block was laid over the eyes of the reconstructed 3D data sets, similar to so-called censor bars in the eye area of standard 2D photos (Fig. 3). For each patient, the size (width, height and depth) and position of a block that could cover both patient eyes completely were determined. Afterwards, all voxels in the area of the block were overwritten with one value. This procedure makes it impossible to reconstruct the original grey values and thus impossible to identify the patient's face. However, the mandible is not affected by this procedure because it is located far below the eyes.

Only this manuscript includes the complete data description, the detailed results and the full data sets of the described CT library. In particular, the complete valid ground truth data and 3D masks are only available within this manuscript.

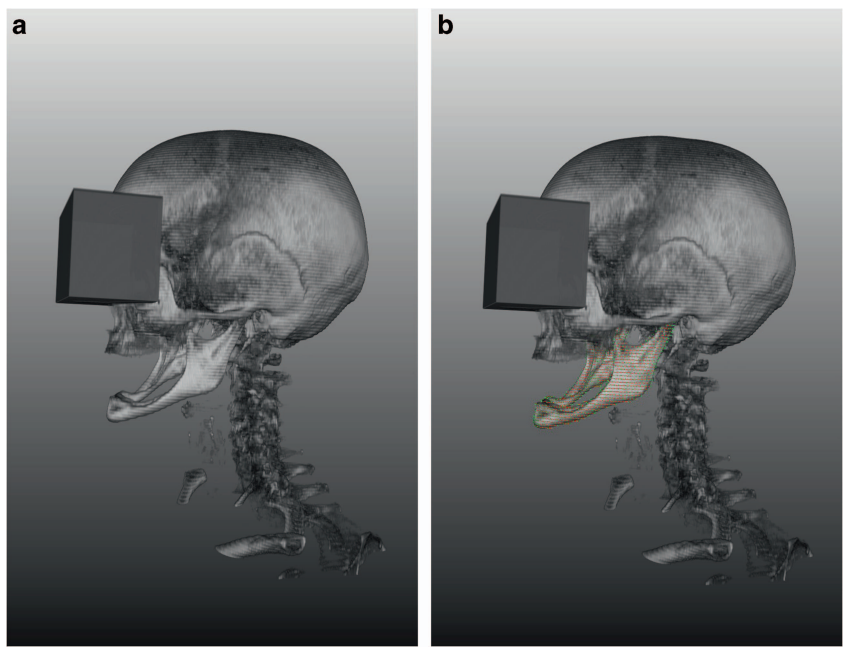

**Figure 3. 3D image data reconstruction.** Each CT image scan consists of multiple slices. In the 3D image reconstructions, a block was laid over the eyes for visual de-identification, similar to so-called censor bars in the eye area of standard 2D photos. (**a**) Original 3D image reconstruction based on multiple slices. (**b**) 3D image reconstruction with the slice-based segmentation mask (3D ground truth mask). *Note:* Fig. 3 *was previously released in part*[17].

## Technical validation

Medical CT scanners used for data collection (Data Citation 1) are regularly subjected to quality control evaluations and are certified medical products, since they are routinely used in daily clinical practice. CT scanners are under the responsibility of a qualified medical physicist.

In routine clinical CT image artefacts, shading can occur near objects of high contrast, such as between bone and soft tissues. The occurrence of these artefacts can be triggered by technical factors but especially by metal materials, such as dental restorations or osteosynthesis plates in the patient[37]. These artefacts can strongly limit diagnostic procedures or image-based segmentation processes, especially if such data sets are used for 3D visualizations of a structure of interest. To obtain valid data sets and avoid such artefacts and limitations, only CT data without teeth and metal materials in the lower jaw were selected[17,21]. Although image-based software programs for the reduction of metal artefacts in the craniomaxillofacial complex exist to potentially improve the diagnostic accuracy of CT images, such as the iterative metal artefact reduction (iMAR) algorithm[38,39], we did not use such image modifications in this work to provide an original data collection. Therefore, a homogeneous data collection without significant artefacts was created to avoid discrepancies between the created ground truth volumes and the scanned anatomical structure caused by artefacts or by missing or damaged slices[17].

The parameters of the CT scanners were set to optimally visualize the mandible and the bone structures of the craniomaxillofacial complex since the CT data were clinically generated for diagnostic reasons. Due to clinical expertise and well-established anatomical cortical structures in the CT scans, manual segmentation was feasible; additional statistical validation of the segmented volumes was performed according to the defined assessment parameters. Segmented 3D image reconstructions were not interpolated between slices to ensure that the resulting 3D models were as accurate as possible (interpolation will probably save time but could introduce some inaccuracies in the visualized results).

Although automatic software segmentation usually provides fast and smooth 3D models from the image slices, we did not use any automatic segmentation software, such as Materialise®, for ground truth model creation. This decision was made to generate ground truth models that are as accurate as possible and created by medical experts without the support of automatic software segmentation. This approach is especially important for atrophic mandibles due to the presence of less calcification. Optimized grey scales and high bone density can reduce these limitations, but a certain so-called image-based smoothed transition remains at the border between hard and soft tissue. In contrast, clinical experts usually interpret anatomical structures more precisely than a standardized algorithm or software program, especially if two experts each segment the same structure independently. However, clinical experts do need more time than a computer algorithm for the segmentation of an anatomical structure. Consequently, the extensive effort and human resources required for this process are probably why accessible manual ground truth segmentations or data collections created directly by clinical experts are rare.

| Mandible No. | Volumes (mm$^3$) | | Ground Truth A vs. Ground Truth B | | Time of segmentation (min) | |
|---|---|---|---|---|---|---|
| | Ground Truth A | Ground Truth B | HD (voxels) | DSC (%) | Ground Truth A | Ground Truth B |
| 1 | 30507.8 | 29413.4 | 3.16 | 94.33 | 36 | 40 |
| 2 | 17333 | 17730.4 | 5.2 | 91.72 | 46 | 40 |
| 3 | 19356.9 | 20067.2 | 3.16 | 92.65 | 38 | 39 |
| 4 | 46506.9 | 47508.8 | 6.32 | 94.66 | 38 | 38 |
| 5 | 39813.6 | 39733 | 3.32 | 93.68 | 37 | 35 |
| 6 | 30861.2 | 31283.1 | 4.12 | 94.48 | 43 | 40 |
| 7 | 45792.7 | 45492.8 | 4.69 | 94.11 | 38 | 42 |
| 8 | 31525.1 | 32288.9 | 2.24 | 94.23 | 36 | 37 |
| 9 | 31368.6 | 31097.9 | 3.32 | 95.44 | 44 | 47 |
| 10 | 19756.3 | 19570.5 | 3.16 | 95.55 | 48 | 51 |

**Table 2. Direct comparison between manual slice-by-slice ground truth segmentations performed by two clinical experts (A, B) for all ten cases. DSC: Dice score coefficient (%), HD: Hausdorff distance (voxels).** Note: Some of the data in Table 2 were previously released in part[17].

To avoid further inaccuracies that probably would have occurred during the manual segmentation process due to inter-observer variability or subjectivity, the ground truth models were statistically tested for validation after the segmentation was completed by the two clinical experts.

All data sets were successfully segmented by the clinical experts without interruptions and were saved as a 3D binary mask. Descriptive and analytical statistical calculations testing the validity of the data are presented in Tables 2–4. The segmentation times of the manual slice-by-slice segmentations were $40.4 \pm 4.43$ min (ground truth A) and $40.9 \pm 4.77$ min (ground truth B) on average (Tables 2 and 3). The mean volume values of the manual slice-by-slice segmentations were $31.28 \pm 10.46$ cm$^3$ (ground truth A) and $31.42 \pm 10.48$ cm$^3$ (ground truth B). On average, the number of voxels in the segmented data sets was $131260 \pm 52273.3$ for ground truth A and $131154.5 \pm 50420.2$ for ground truth B. The overlap agreement between the two manual segmentations (A, B) yielded a Dice score of $94.09 \pm 1.17\%$. The calculated HD between the two manual segmentations (A, B) was $3.87 \pm 1.21$ voxel units. These values both represent a close agreement between the sets of 3D models, since the Dice score values are high, showing a high percentage of agreement in the 3D models, and the HD was low, showing small voxel differences between the sets of 3D models. The calculated differences between the ground truth segmentations (A, B) were not significant for the assessment parameters volume and voxels ($p > 0.05$) (Fig. 4a and c). Furthermore, the product-moment correlation coefficient (Pearson) in the regression models for volume and voxels was close to the value one and was not below 0.99 ($r > 0.99$) when the two segmentation groups were compared (Fig. 4b and d).

The measurement values of the segmentation volumes and voxels in the regression models were localized closely along the regression lines (Fig. 4b and d). The resulting boxplots are similar in terms of the volume and voxel values between the segmentations made by clinical experts A and B (Fig. 4a and c). This comparison is also valid for the parameters DSC and HD, as shown in Fig. 5a and b. Both parameters were similar in a comparison between manual ground truth segmentations A and B (Tables 3 and 4).

High DSC values, low HD values, strong positive correlations and measurement values close to the regression line were achieved between the compared segmentation models without significant variability or inaccuracy, which demonstrates strong accordance between the segmentation models and the valid ground truth data.

For a visual assessment, Fig. 6 presents the overlap of two manual ground truth segmentations (grey and turquoise) using a 3D visualization of one case. Since the ground truth schemes were created manually according to the number of image slices without modifying or interpolating slices, a clear tomography-based 3D visualization is achieved in the reconstructed 3D segmentation model. The slice-by-slice ground truth segmentation is truly visualized as a stepwise surface countering, meaning each visualized step defines one slice.

However, for a more quantitative assessment of the agreement between the 3D models, interested users could use a program like CloudCompare (https://www.danielgm.net/cc/) to compare pairs of models. This process is similar to a study that compared 3D models of fossils produced with different techniques[40].

| | Volumes of Mandibles (cm³) | | Ground Truth A vs. Ground Truth B | | Time of segmentation (min) | |
|---|---|---|---|---|---|---|
| | Ground truth A | Ground truth B | HD (voxels) | DSC (%) | Ground Truth A | Ground Truth B |
| Min | 17.33 | 17.73 | 2.24 | 91.72 | 36 | 35 |
| Max | 46.51 | 47.51 | 6.32 | 95.55 | 48 | 51 |
| $\mu \pm \sigma$ | 31.28 ± 10.46 | 31.42 ± 10.48 | 3.87 ± 1.21 | 94.09 ± 1.17 | 40.4 ± 4.43 | 40.9 ± 4.77 |

**Table 3. Summary of manual vs. manual (Ground truth A, B) segmentation results, presenting minimum (Min) values, maximum (Max) values, mean (μ) values and standard deviations (σ) for mandibles.** DSC: Dice score coefficient (%), HD: Hausdorff distance (voxels). Note: Some of the data in Table 3 were previously released in part[17].

| Mandible No. | Voxels | |
|---|---|---|
| | Ground Truth A | Ground Truth B |
| 1 | 166749 | 160767 |
| 2 | 118277 | 120989 |
| 3 | 54887 | 56901 |
| 4 | 84897 | 86726 |
| 5 | 153211 | 152901 |
| 6 | 96836 | 98160 |
| 7 | 211925 | 210537 |
| 8 | 77436 | 79312 |
| 9 | 188773 | 187144 |
| 10 | 159609 | 158108 |
| Min | 54887 | 56901 |
| Max | 211925 | 210537 |
| $\mu \pm \sigma$ | 131260 ± 52273.3 | 131154.5 ± 50420.2 |

**Table 4. Direct comparison of the number of voxels for the manual segmentations (Ground Truth A, B) for all ten cases, presenting minimum (Min) values, maximum (Max) values, mean (μ) values and standard deviations (σ).** Note: Some of the data in Table 4 were previously released in part[17].

Although the use of many image-based segmentation algorithms has been reported for the craniomaxillofacial complex[41–45] and many interactive image-based segmentation approaches can be found in the literature, as overviewed by Zhao and Xie[46], for example, the ability of available valid medical ground truth data of complete bone structures to reproducibly investigate or test image-processing techniques or segmentation algorithms is limited[17,21]. Accessible medical databases are limited, and valid medical ground truth databases for the evaluation of algorithms are rare and usually comprise only a few images. Inaccuracy or invalidity of medical ground truth data and image-based artefacts also limit the creation of such databases. This issue is especially relevant for CT data sets of the craniomaxillofacial complex.

If available, medical ground truth models are often incomplete or include so-called faulty slices due to skipped slice series or radiological artefacts during the creation of these models. Especially for the craniomaxillofacial complex, such incomplete ground truth models are not relevant because parts of the anatomical structures are missing. For example, artefacts from dental restorations, which can be present in multiple locations in a single patient, usually include between 20–30 slices depending on the slice thickness of the data set (Fig. 7). Additionally, 3D medical ground truth models are strongly associated with certain inaccuracies due to inter- and intra-observer deviation in the segmentation process during ground truth creation, which can lead to subjectivity or invalid models[21].

However, this contribution provides a description of an accessible and unique data collection of 10 valid ground truth segmentation models of the complete mandible originating from routine clinical practice without artefacts or faulty slices (Data Citation 1). The ground truth data were created by medical specialists[17,21] and statistically tested for validation to avoid invalid models. These data provide resources for conducting serial image studies of the human mandible, evaluating segmentation algorithms and developing adequate image tools for clinical or scientific use. The accuracy and functional stability of segmentation algorithms can be further tested for reproducibility in comparison to other open-source based image tools or CE-certified medical software products, such as MIMICS (https://biomedical.

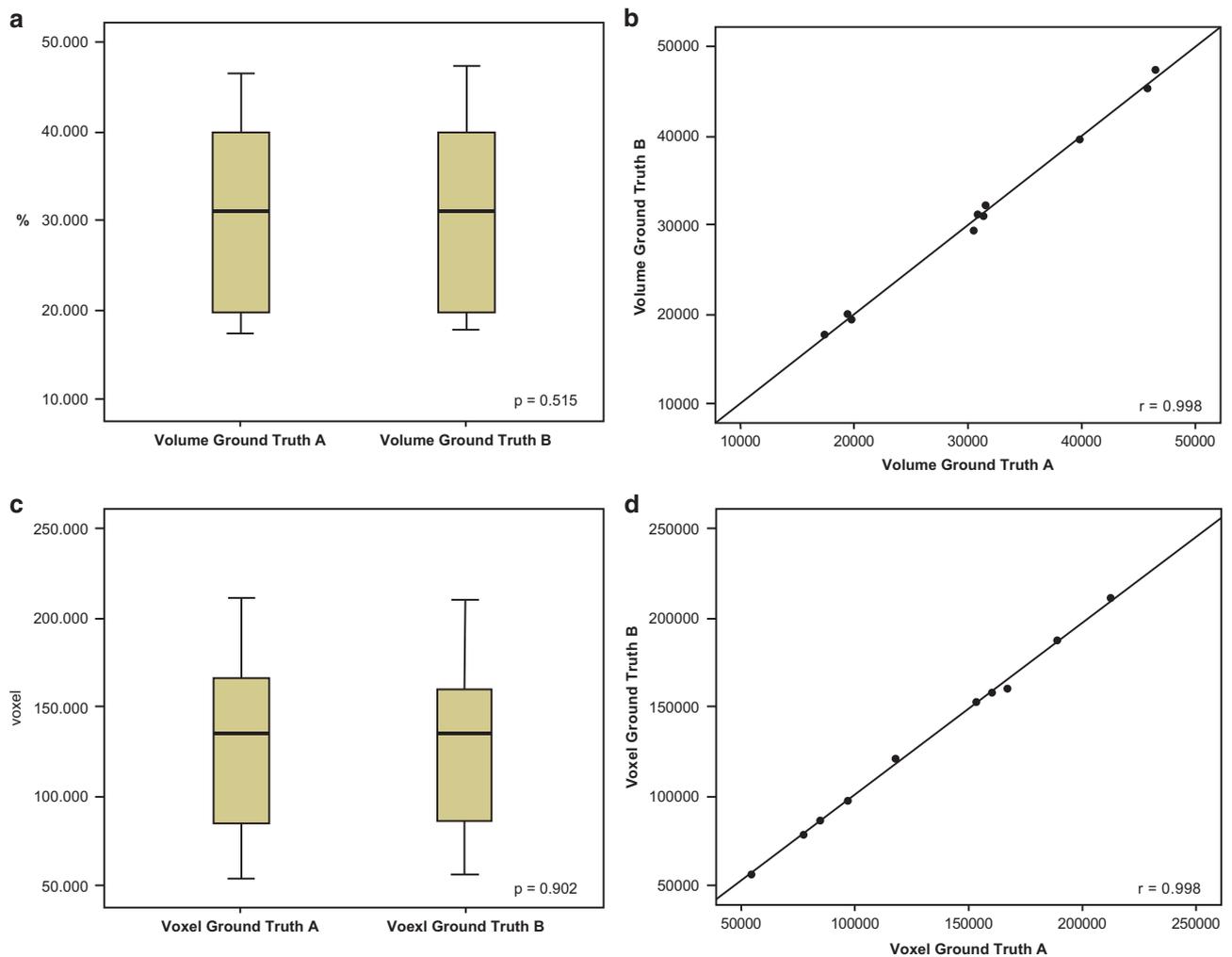

Figure 4. **Data validity**. Paired t-tests (p), Pearson's product-moment correlation coefficient (r) and regression analysis, including boxplots and regression lines through the origin, were used to test the data validity. Strong positive correlations and measurement values close to the regression line were achieved between the compared segmentation models without significant variability or inaccuracy, which proves the validity of the ground truth data. (**a**) The calculated difference between the ground truth segmentations (A, B) was not significant for the assessment parameter volume ($p > 0.05$), which is graphically shown in the boxplots, which are similar between segmentations A and B. (**b**) The measured volume values were localized closely along the regression line; thus, the product-moment correlation coefficient (Pearson) of the regression model volume was close to the value one, not below 0.99 ($r > 0.99$). (**c**) The calculated difference between the ground truth segmentations (A, B) was not significant for the assessment parameter voxels ($p > 0.05$), which is graphically shown in the boxplots, which are similar between segmentations A and B. (**d**) The measured voxel values were localized closely along the regression line; thus, the product-moment correlation coefficient (Pearson) in the regression model voxel was close to the value one and was not below 0.99 ($r > 0.99$). *Note: Some of the data in* Figure 4 *were previously released in part*[17].

materialise.com/mimics), with the provided data collection (Data Citation 1), clinically and/or scientifically.

The presented clinical CT data collection (Data Citation 1) may be of high relevance because image-based software programs have gained importance in various medical fields in recent years[47], and the evaluation of image-based software segmentation is therefore a central point of interest in related computer-assisted workflow procedures both clinically and scientifically[7,8,48–50]. Additionally, image-based software segmentation has rarely been evaluated for the complete mandible, and an available 3D ground truth data collection of the complete human lower jaw (Data Citation 1) has been missing so far.

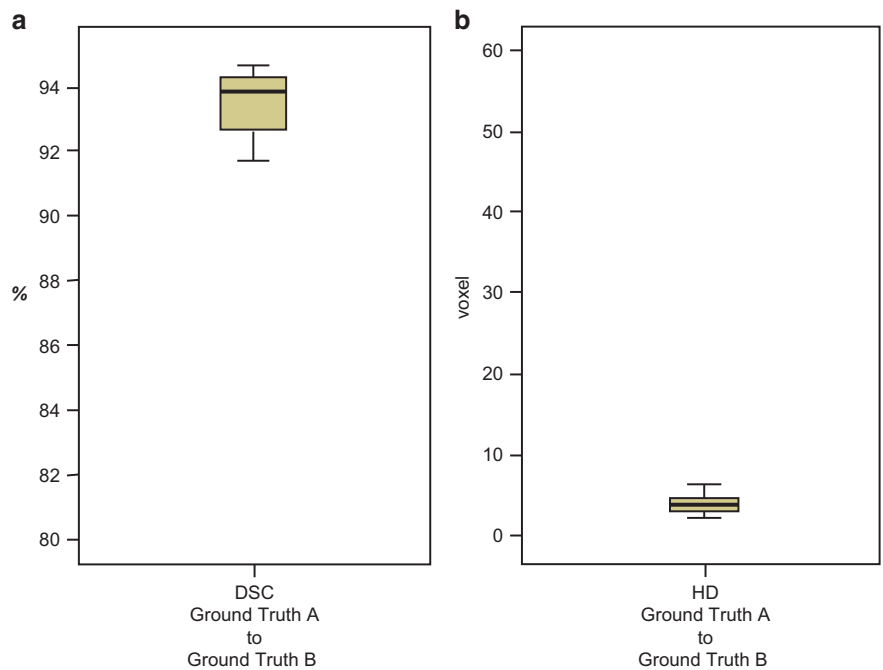

Figure 5. **Comparison of ground truth data**. The accuracy overlay between the ground truth segmentations was compared using DSC (%) and HD (voxel) values. The boxplots of DSC (%) and HD (voxels) showed high DSC and low HD values, which proves the strong accordance between the segmentation models. *Note: Some of the data in* Figure 5 *were previously released in part*[17].

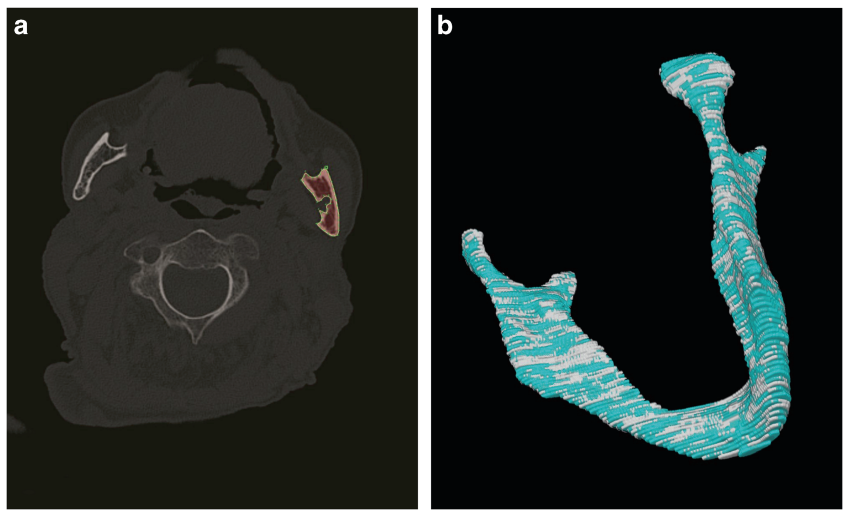

Figure 6. **Visual segmentation assessment**. Each case was segmented twice by two clinical experts, A and B (ground truth segmentation). (**a**) The segmentation of a bone structure (red field) is shown in 2D. (**b**) The overlap of two manual ground truth segmentations A and B (grey and turquoise) of one case is shown in a 3D visualization. Since the ground truth schemes were created manually according to the number of image slices, a clear tomography-based 3D visualization is achieved in the reconstructed 3D segmentation model without image interpolation. *Note:* (**b**) *was previously released in part*[17].

The main research highlights of the presented work are as follows:

- Manual slice-by-slice segmentations of mandibles were performed by clinical experts[17,21] to obtain ground truth data for the lower jawbone boundaries and estimates of rater variability.

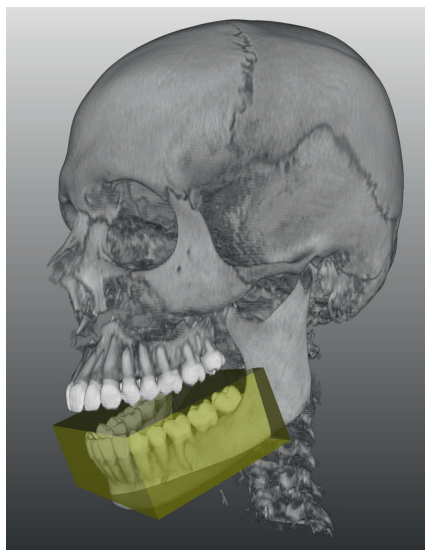

Figure 7. **Artefact effects of mandible CT data sets**. Degree of potentially affected slices due to artefacts caused by dental restorations in the lower jaw (yellow). These artefacts frequently occur in CT data sets with teeth due to metal dental materials. The artefacts can lead to incomplete segmentation models and cause missing or faulty slices or missing visualized anatomical parts in the 3D reconstruction that alter the original data set. This issue is especially important when fully automatic algorithmic segmentation processes are used. Therefore, only CT data sets without teeth were used in this study.

- Quality evaluation of the ground truth segmentations and comparisons were performed by calculating DSCs[30,31] and HDs[32] and performing statistical analysis to directly compare the volume, voxel, DSC and HD values[17] of the created ground truth models.
- A unique clinical CT image data set collection of the mandible, fully segmented ground truth data and 3D masks of the same anatomical structure are provided and have been statistically tested for validity for reproducible use by others.
- The CT image data set collection, ground truth data and 3D masks were originally created without artefacts or missing slices to provide complete data sets.
- The provided data can be used for the further assessment or development of varying segmentation approaches, for reproducible investigation of image-processing tools or for individual research purposes (please see the Acknowledgements section for more information).

Despite the extensive data presented in this work, there are several areas for future work, such as the addition of more data over time and the generation of more manual ground segmentations (by clinical experts from other institutions) to create an expanding clinical CT data platform for the investigation of previously and newly developed image-processing tools or software tools.

## Usage notes

The anonymized CT scan data collection (Data Citation 1) is stored as single .nrrd files, e.g., "pat1.nrrd", and can be processed with common and free medical platforms, such as MeVisLab (https://www.mevislab.de/)[51], 3D Slicer (https://www.slicer.org)[33] and MITK (www.mitk.org)[52], and tools such as MATLAB (https://www.mathworks.com). 3D models (Figs 3 and 8) can be generated and exported in various formats by using the software packages mentioned above. In addition to these file formats, we provide CT scans as stacked TIFFs and manual segmentations in the standard 3D file format STL to increase accessibility for more potential users[53]. The .nrrd files were converted to stacked TIFFs with Slicer 4.6.2[54], and the manual segmentation masks were converted to STL files with MeVisLab 2.7[55] and the WEMIsoSurface module. The corresponding mandible contours (ground truth segmentations) will be provided as CSO files, whereby an "a" stands for the segmentations of physician A and a "b" stands for the segmentations of physician B. Moreover, we encoded the segmentation times in the filenames, for example, "Pat 1a (36 min 15 sec).cso" or "Pat 1b (39 min 43 s)". The CSO format was used in this work to save the individual outlines of the clinical experts. These CSO data are not altered in any way. In addition, we will also provide the manual segmentations as voxelized 3D masks in .nrrd format (note: in contrast to the outlines, these masks are filled). Thereby, the masks fit to the corresponding patient file, for example "mask1.nrrd" belongs to "pat1.nrrd" and so on (note: we used the same file name convention described for the CSOs, e.g., "Pat 1a (36 min 15 sec) Manual 3D Mask.nrrd"). The corresponding files are in the same coordination system, which means they can be loaded and overlaid without any registration (Fig. 3). The CT data collection (Data Citation 1) is stored as .nrrd files and can be used to test and compare conventional freely available image-based algorithms or commercial financially available software processing tools for the mandible for clinical or scientific use. The quality and accuracy of software programs can be evaluated in comparison to the appropriate ground truth data provided as CSO files using, for example, DSC and HD calculations, as described in this manuscript. Therefore, the ground truth data set described in this work can be used as a control group in comparative assessments of image-based computer programs and segmentation algorithms that are tested on CT data for the mandible. Additionally, 3D models can be produced when software programs are tested on this data collection (Data Citation 1) and superimposed on the ground truth CSO data for a visual assessment. 3D image reconstructions of structures of interest can be further generated for quantitative and visual testing of morphological variability.

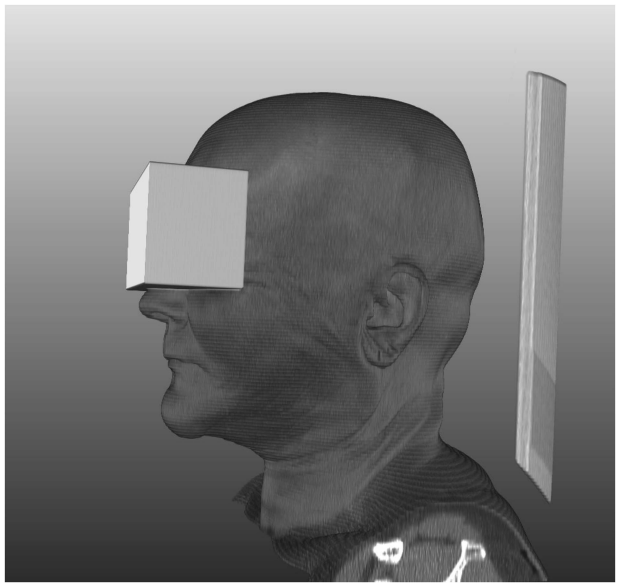

**Figure 8. 3D models.** Soft and/or hard tissue 3D models can be generated and exported in various formats by using free software packages. For visual de-identification, a block was laid over the eyes, similar to so-called censor bars in the eye area of standard 2D photos.

Finally, we should mention that we did not perform any postprocessing of the masks, e.g., 3D surface smoothing, because we did not want to alter the manual slice-by-slice segmentations, which would alter the ground truth.

For research purposes, the data can freely be downloaded, but we kindly ask investigators to cite our work. The data provided within this work are free to share — copy and redistribute the material in any medium or format. Furthermore, the data are free to adapt — remix, transform, and build upon the material for any purpose. The data within this work are licensed under a Creative Commons Attribution 4.0 International License (CC BY 4.0) (https://creativecommons.org/licenses/by/4.0/).

The data files attached to this manuscript can be opened in ImageJ with good results and can be displayed at full quality. The files can also be opened in Irfanview, but the data do not display at full quality.

### Data Citation

1. Wallner, J. & Egger, J. *Figshare* https://doi.org/10.6084/m9.figshare.6167726.v5 (2018).

### Acknowledgements

This investigation was approved by the internal review board (IRB) of the Medical University of Graz, Austria (IRB: EK-29-143 ex 16/17). This work received funding from the Austrian Scientific Fund (FWF-KLIF): "Enfaced: Virtual and Augmented Reality Training and Navigation Module for 3D-Printed Facial Defect Reconstructions" (KLI-678-B31; PIs: Jürgen Wallner and Jan Egger), the TU Graz Lead Project (Mechanics, Modeling and Simulation of Aortic Dissection) and CAMed (COMET K-Project 871132), which is funded by the Austrian Federal Ministry of Transport, Innovation and Technology (BMVIT) and the Austrian Federal Ministry for Digital and Economic Affairs (BMDW) and the Styrian Business Promotion Agency (SFG). Some of the data in this collection were partly used in previously published works[17,22,23]. In these publications, the data were used as testing data for single algorithms or deep learning networks. Some CT data sets were used to test a deep learning network[22] in the mandible for fully automatic network-based segmentation and to test the accuracy of the licence-free open-source algorithm GrowCut (https://www.growcut.com) in the mandible[17]. In this investigation[17], parts of the CT data that were used for algorithmic testing were initially uploaded as a figshare repository for reproducibility reasons because of the journal's requirements for publication. The overall HD and DSC results presented in this publication were partly presented within a conference paper[21]. Further, one CT dataset was used in .stl (standard triangle language) file format as a surface model without segmentation to evaluate a software tool for computer-aided planning of miniplate positioning for oral and maxillofacial surgery[23]. However, only this manuscript includes the complete data description, the detailed results and the full data sets of the described CT library. In particular, the complete valid ground truth data and 3D masks are only presented within this manuscript.

### Author Contributions

Conceived and designed the experiments: J.W. and J.E. Performed the experiments: J.W. and J.E. Analysed the data: J.W., I.M., and J.E. Contributed reagents/materials/analysis tools: J.W., I.M., and J.E. Wrote the paper: J.W. and J.E.

### Additional Information

**Competing interests**: The authors declare no competing interests.

**How to cite this article**: Wallner, J. *et al*. Computed tomography data collection of the complete human mandible and valid clinical ground truth models. *Sci. Data.* 6:190003 https://doi.org/10.1038/sdata.2019.3 (2019).

**Publisher's note**: Springer Nature remains neutral with regard to jurisdictional claims in published maps and institutional affiliations.